\def\year{2022}\relax
\documentclass[letterpaper]{article} 
\usepackage{aaai22}  
\usepackage{times}  
\usepackage{helvet}  
\usepackage{courier}  
\usepackage[hyphens]{url}  
\usepackage{graphicx} 
\usepackage[round]{natbib}
\usepackage{caption} 
\DeclareCaptionStyle{ruled}{labelfont=normalfont,labelsep=colon,strut=off} 
\usepackage{algorithm}
\usepackage{algorithmic}

\usepackage{newfloat}
\usepackage{listings}
\floatstyle{ruled}
\newfloat{listing}{tb}{lst}{}
\floatname{listing}{Listing}
\usepackage{bm}
\usepackage{amsthm}
\usepackage{amssymb}
\usepackage{amsmath}
\usepackage{mathtools}
\usepackage{pgfplots}

\usepackage{tikz}
\usetikzlibrary{
  backgrounds,
  intersections,
  shapes
}
\usepgfplotslibrary{groupplots}
\usepgfplotslibrary{fillbetween}
\pgfplotsset{
compat=1.11,
    legend image code/.code={
        \draw[mark repeat=2,mark phase=2]
        plot coordinates {
            (0cm,0cm)
            (0.15cm,0cm)
            (0.3cm,0cm)
        };%
    }
}

\usepackage{pifont}
\newcommand{\cmark}{{\color{green} \ding{51}}}
\newcommand{\xmark}{{\color{red} \ding{55}}}

\usepackage{enumitem}
\usepackage{dsfont}

\newcommand\markit[1]{\tikz[scale=1.4]\pgfuseplotmark{#1};}
\nocopyright
\usepackage[hidelinks,breaklinks]{hyperref}
\usepackage{multicol}
\usepackage{balance}
\usepackage[resetlabels]{multibib}
\newcites{supp}{References}

\title{Fitting Large Mixture Models Using Stochastic Component Selection\footnote{The authors acknowledge the support of the OP VVV funded project CZ.02.1.01/0.0/0.0/16\_019/0000765 ``Research Center for Informatics''.}}
\author{
    Milan Pape\v{z}, Tom\'{a}\v{s} Pevn\'{y}, V\'{a}clav \v{S}m\'{i}dl
}
\affiliations{
    Artifcial Intelligence Center\\
    Czech Technical University, Prague, Czech Republic\\
    $\lbrace$papezmil, pevnytom, smidlva1$\rbrace$@fel.cvut.cz
}

\usepackage{bibentry}

\begin{document}

\maketitle

\begin{abstract}
Traditional methods for unsupervised learning of finite mixture models require to evaluate the likelihood of all components of the mixture. This becomes computationally prohibitive when the number of components is large, as it is, for example, in the sum-product (transform) networks. Therefore, we propose to apply a combination of the expectation maximization and the Metropolis-Hastings algorithm to evaluate only a small number of, stochastically sampled, components, thus substantially reducing the computational cost. The Markov chain of component assignments is sequentially generated across the algorithm's iterations, having a non-stationary target distribution whose parameters vary via a gradient-descent scheme. We put emphasis on generality of our method, equipping it with the ability to train both shallow and deep mixture models which involve complex, and possibly nonlinear, transformations. The performance of our method is illustrated in a variety of synthetic and real-data contexts, considering deep models, such as mixtures of normalizing flows and sum-product (transform) networks.
\end{abstract}

\section{Introduction}
Finite mixture models \cite{mclachlan2019finite} constitute a fundamental class of density estimation models. They are formed by a weighted sum of probability distributions---here referred to as \emph{components}---to cluster $N$ unlabelled datapoints into $K$ categories. Their training is challenging in applications involving a large number of categories, $K$, such as face recognition \cite{otto2017clustering}, astronomical imaging \cite{welton2013mr}, natural language processing \cite{nayak2014clustering} and DNA data storage \cite{rashtchian2017clustering}. The maximum likelihood techniques train these models by optimizing either (i) the marginal likelihood via gradient-descent \cite{redner1984mixture} or (ii) the evidence lower bound via variational methods \cite{blei2017variational}, including the expectation-maximization (EM) \cite{dempster1977maximum}. The dependence structure among variational distributions then ranges from fully independent (mean-field) \cite{humphreys2000approximate} to fully dependent \cite{kucukelbir2017automatic}. The sampling-based methods target the posterior distribution using sequential Monte Carlo \cite{chopin2002sequential} or Markov chain Monte Carlo (MCMC) \cite{robert2013monte}, e.g.\ via the Gibbs \cite{lavine1992bayesian} or Metropolis-Hastings sampling \cite{marin2005bayesian}. The computational cost of such methods scales with $\mathcal{O}(TDNK)$ operations, where $N$ and $K$ are defined above, $T$ is the number of iterations, and $D$ is the dimension of data.

However, all the aforementioned techniques---including those which reduce the computational cost via the first three factors in $\mathcal{O}(TDNK)$, see Section \ref{sec:related_work}---evaluate all $K$ components. This is demanding for large models, and the problem is even more severe for mixtures utilizing intricate structures, such as neural networks \cite{greff2017neural,monnier2020deep}, Gaussian processes \cite{wu2019effective}, normalizing flows \cite{pires2020variational}; or deep mixtures, such as sum-product (transform) networks \cite{peharz2020random,pevny2020sum} and deep Gaussian mixture models \cite{viroli2019deep}. In spite of this, a little attention has been paid to the design of algorithms that do not evaluate all $K$ components. The notable exceptions are the sparse EM algorithm \cite{hughes2016fast} and the truncated EM algorithm \cite{forster2018can}, see Table \ref{tab:algorithms}. Moreover, the methods are mostly tailored for a fixed class of mixture models, e.g.\ the Gaussian mixture models (GMMs).

\begin{table*}
    \small
    \centering
    \begin{tabular}{lcccc}
    \hline
        Algorithm / Feature & $B<N$ datapoints & $M<K$ statistics & $M<K$ likelihoods & deep models\\
    \hline
        EM \cite{dempster1977maximum} & \xmark & \xmark & \xmark & \xmark \\
        SAEM \cite{nguyen2020mini} & \cmark & \xmark & \xmark & \xmark \\
        SSAEM \cite{hughes2016fast} & \cmark & \cmark & \xmark & \xmark \\
        TSAEM \cite{forster2018can} & \cmark & \cmark & \cmark & \xmark \\
        MCSAEM \cite{allassonniere2021new} & \cmark & \cmark & \xmark & \xmark \\
        MHSAEM (ours) & \cmark & \cmark & \cmark & \cmark \\
    \hline
    \end{tabular}
    \caption{The computational features of various EM algorithms (Section \ref{sec:related_work}). We compare whether the methods (i) perform the computations with a reduced number of data (minibatching), (ii) update a lower number of statistics, (iii) make less evaluations of the conditional likelihood, and (iv) are suitable for training of deep models. Here, EM, SA, S, T, MC and MH stand for expectation-maximization, stochastic approximation, sparse, truncated, Monte Carlo and Metropolis-Hastings, respectively.
    }
    \label{tab:algorithms}
\end{table*}

In this paper, we make the following contributions:
\begin{itemize}
    \item We instantiate the MCMC stochastic approximation EM (MCMCSAEM) framework \cite{kuhn2004coupling,kuhn2020properties} in the context of finite mixture models. This may seem wasteful, since the expectation in the EM objective function is analytically tractable, whereas the MCMCSAEM framework is notoriously applied when the expectation is intractable. However, going against this convention allows us to evaluate less components in mixture models, substantially reducing the computational cost of their training.
    \item We design our method to enable the maximization of generic EM objectives via a gradient-descent type of stochastic approximation, making it suitable for deep mixture models. This enhances the MCMCSAEM framework which has been almost exclusively used in cases admitting the maximization under a closed-form solution.
    \item We apply our method to GMMs and their non-trivial generalizations: sum-product-transform networks (SPTNs) and mixtures of real-valued non-volume preserving (real NVP) flows \cite{dinh2017density}, reaching $\text{100}\times$ speed-up compared to the state-of-the-art.
\end{itemize}

\section{Expectation maximization}
\subsection{The EM algorithm}
The EM algorithm \cite{dempster1977maximum} seeks the unknown parameters, $\theta\in\Theta$, maximizing the marginal (incomplete-data) log-likelihood in latent data models,
\begin{equation}
    \mathcal{L}(\theta)
    \coloneqq
    \log p_{\theta}(\mathbf{x})
    =
    \log\int p_{\theta}(\mathbf{x},\mathbf{z})d\mathbf{z}
    ,
    \label{eq:incomplete_data_likelihood_plain}
\end{equation}
where $\mathbf{x}\coloneqq(x_i)^N_{i=1}$ are observed (known) variables, $x\in\mathsf{X}$; $\mathbf{z}\coloneqq(z_i)^N_{i=1}$ are latent (unknown) variables, $z\in\mathsf{Z}$; and $p_{\theta}(\mathbf{x},\mathbf{z})$ is the joint (complete-data) likelihood.

The EM algorithm addresses this task indirectly, i.e.\ by optimizing the evidence lower bound (ELBO),
\begin{equation}
    \mathcal{L}(\theta)\geq\mathcal{Q}(\theta,\hat{\theta})+\mathcal{H}(\hat{\theta})
    \coloneqq\text{ELBO}(\hat{\theta}),
    \label{eq:elbo}
\end{equation}
where $\mathcal{H}(\hat{\theta})\coloneqq-\mathsf{E}_{p_{\hat{\theta}}(\mathbf{z}|\mathbf{x})}[\log p_{\hat{\theta}}(\mathbf{z}|\mathbf{x})]$ is the differential entropy at an estimate, $\hat{\theta}\in\Theta$, and
\begin{equation}
    \mathcal{Q}(\theta,\hat{\theta})
    \coloneqq
    \mathsf{E}_{p_{\hat{\theta}}(\mathbf{z}|\mathbf{x})}[\log p_{\theta}(\mathbf{z},\mathbf{x})]
    \label{eq:intermediate_quantity}
\end{equation}
is the EM objective function. Here, $p_{\theta}(\mathbf{z}|\mathbf{x})$ is the posterior distribution over $\mathbf{z}$. Given an initial value, $\theta_1$, the algorithm produces a sequence of estimates, $(\theta_t)^{T}_{t=1}$, by alternating between the expectation (E) and maximization (M) steps,
\begin{align}
    \text{E-step:} &\hspace{5pt} \mathcal{Q}_{t}(\theta)
    ,
    \label{eq:generic_e_step}
    \\
    \text{M-step:} &\hspace{5pt} \theta_{t+1}\coloneqq\underset{\theta\in\Theta}{\operatorname{arg\hspace{1pt}max}}\,\mathcal{Q}_{t}(\theta)
    ,
    \label{eq:m_step}
\end{align}
where $\mathcal{Q}_{t}(\theta)\coloneqq\mathcal{Q}(\theta,\theta_t)$. The sequence is guaranteed to monotonically tighten the ELBO, \eqref{eq:elbo}, arriving at a local optimum of \eqref{eq:incomplete_data_likelihood_plain} under mild regularity assumptions~\cite{wu1983convergence}.

\subsection{The MCMCSAEM algorithm}\label{sec:mcmcsaem}
The E-step \eqref{eq:generic_e_step} is analytically intractable in many applications. The MCEM algorithm \cite{wei1990monte} addresses this problem by approximating \eqref{eq:intermediate_quantity} via the MC average, $\bar{\mathcal{Q}}_{t}(\theta)\coloneqq\frac{1}{M}\sum^M_{j=1}\log p_\theta(\mathbf{x},\mathbf{z}_t^j)$, where the samples, $(\mathbf{z}_t^j)^M_{j=1}$, are drawn from $p_{\theta_t}(\mathbf{z}|\mathbf{x})$. However, this algorithm requires high values of $M$ to converge \cite{fort2003convergence}, and the samples are wastefully discarded at each iteration, $t$. The stochastic approximation resolves this issue by reusing the samples in $\bar{\mathcal{Q}}_{t}$ over the iterations as follows:
\begin{equation}
    \textstyle\widehat{\mathcal{Q}}_{t}(\theta)\coloneqq\widehat{\mathcal{Q}}_{t-1}(\theta)+\gamma_t\big(\bar{\mathcal{Q}}_{t}-\widehat{\mathcal{Q}}_{t-1}(\theta)\big)
    ,
    \label{eq:stochastic_approximation_generic}
\end{equation}
where the step-size, $\gamma_t$, satisfies the constraints \cite{robbins1951stochastic}, $\gamma_t\in[0, 1]$, $\sum_{t\geq 1}\gamma_t=\infty$, $\sum_{t\geq 1}\gamma^2_t<\infty$.

The normalizing factor of $p_\theta(\mathbf{z}|\mathbf{x})$ is often intractable, preventing direct sampling from this posterior. MCMC obviates this difficulty by simulating a Markov chain, $(\mathbf{z}_t)^T_{t=1}$, from a transition kernel, $\mathbf{z}_t\sim P_\theta(\mathbf{z}_{t-1}, \cdot)$, which leaves $p_\theta(\mathbf{z}|\mathbf{x})$ as its unique stationary distribution, given a fixed~$\theta$.

The MCMCSAEM algorithm \cite{delyon1999convergence,kuhn2004coupling,kuhn2020properties} approximates the E-step \eqref{eq:generic_e_step} by combining the MCMC simulation (S) and the stochastic approximation (SA) in \eqref{eq:stochastic_approximation_generic},
\begin{align}
    \text{S-step:} &\hspace{5pt} \mathbf{z}^j_t\sim P_{\theta_t}(\mathbf{z}^{j-1}_t, \cdot)
    ,\hspace{15pt}j\in(1,\ldots,M),
    \label{eq:s_step_generic}
    \\
    \text{SA-step:} &\hspace{5pt} \widehat{\mathcal{Q}}_{t}(\theta)
    ,
    \label{eq:sa_step_generic}
    \\
    \text{M-step:} &\hspace{5pt} \theta_{t+1}\coloneqq\underset{\theta\in\Theta}{\operatorname{arg\hspace{1pt}max}}\,\widehat{\mathcal{Q}}_{t}(\theta)
    \label{eq:m_step_generic}
    .
\end{align}
This algorithm sets $\mathbf{z}^0_t\coloneqq \mathbf{z}^M_{t-1}$ at each $t$ and produces the chain $(\mathbf{z}^1_1,\ldots,\mathbf{z}^M_1,\ldots,\mathbf{z}^1_T,\ldots,\mathbf{z}^M_T)$ of length $MT$, initializing \eqref{eq:s_step_generic} with $\mathbf{z}^0_1$. The samples generated during the initial iterations are usually discarded due to their high correlation, which is often referred to as the burn-in \cite{robert2013monte}. In the MCMCSAEM algorithm, the initial samples do not have to be discarded, since they are sequentially forgotten via the step-size, $\gamma_t$, (forgetting factor) in \eqref{eq:stochastic_approximation_generic}, i.e.\ a specific form of the sequence, $(\gamma_t)^T_{t=1}$, handles the burn-in.

Even for $M=1$, the chain, $(\mathbf{z}_t)^T_{t=1}$, is non-homogeneous, since $\theta_t$ varies over the iterations, and the target distribution, $p_{\theta_t}(\mathbf{z}|\mathbf{x})$, is thus non-stationary. The convergence of $(\theta_t)^{T}_{t=1}$ towards a stationary point of $\mathcal{L}(\theta)$ was proven under the restriction of $p_\theta(\mathbf{x},\mathbf{z})$ belonging to the exponential family \cite{kuhn2004coupling}, where \eqref{eq:stochastic_approximation_generic} reduces to closed-form updates of finite-dimensional, sufficient statistics.

\section{Problem formulation}\label{sec:problem_formulation}
A finite mixture model characterizes the relation between $x\in\mathsf{X}\subseteq\mathbb{R}^{D}$ and $z\in\mathsf{Z}\coloneqq\lbrace{1,\ldots, K\rbrace}$ as follows:
\begin{equation}
    p_{\theta}(x)
    =
    \sum^K_{k=1}p_{\eta_k}(x|z=k)p_{\pi_k}(z=k)
    ,
    \label{eq:incomplete_data_likelihood_generic}
\end{equation}
where $\theta\coloneqq(\pi_1,\eta_1,\ldots,\pi_K,\eta_K)$ are unknown parameters. $\eta_z$ are the parameters of the conditional likelihood, $p_{\eta_z}(x|z)$, and $\pi_z$ is the weight parameterizing the prior, $p_{\pi_z}(z)=\pi_z$, such that $0\leq\pi_k\leq 1$ for each $k\in\mathsf{Z}$ and $\sum_{k=1}^K\pi_k=1$.


Given independent and identically distributed data, $\mathbf{x}$, our aim is to find the parameters maximizing \eqref{eq:incomplete_data_likelihood_plain} given by
\begin{equation}
    \mathcal{L}(\theta)
    =
    \sum^N_{i=1}\log\sum^K_{k=1}p_{\eta_k}(x_i|z_i=k)p_{\pi_k}(z_i=k)
    .
    \label{eq:incomplete_data_likelihood_specific}
\end{equation}
For \eqref{eq:incomplete_data_likelihood_generic}, the integration in \eqref{eq:incomplete_data_likelihood_plain} becomes the summation, which is analytically tractable for all forms of $p_{\eta_z}(x|z)$. Indeed, we consider $p_{\eta_z}(x|z)$ to belong to an arbitrary family of $\eta_z$-differentiable probability distributions. However, for high $K$, the marginalization in \eqref{eq:incomplete_data_likelihood_specific} is computationally costly, which renders the optimization objective presumably intractable. We would like to design an algorithm requiring only $M<K$ evaluations of $p_{\eta_z}(x|z)$ at each $t$.

\section{The generalized MHSAEM algorithm}
The application of the MCMCSAEM framework is notoriously motivated by analytical intractability of the E-step. We go against this convention, and use it to reduce the computational cost of the EM algorithm in the context of finite mixture models, where the E-step---the finite sum expected value---is always tractable. Moreover, the MCMCSAEM algorithm involves a closed-form solution of the M-step. We release this assumption by allowing direct, gradient-based optimization of the EM objective.

\subsection{E-step}
The computational cost of the EM algorithm scales with $\mathcal{O}(TDNK)$, since \eqref{eq:generic_e_step} factorizes as follows:
\begin{equation}
    \mathcal{Q}_t(\theta)=\sum^N_{i=1}\sum^K_{k=1}p_{\theta_t}(z_i=k|x_i)\log p_{\theta}(z_i=k,x_i)
    ,
    \nonumber
\end{equation}
which contains $KN$, $D$-dependent, summands that are re-computed at each $t\in(1,\ldots,T)$. Indeed, the marginal factor, $p_{\pi_z}(z)$, of $p_\theta(x,z)$ is just the cheap categorical distribution; however, the conditional factor, $p_{\eta_z}(x|z)$, typically involves high-dimensional operations (e.g.\ the inversion of $D\!\times\!D$-dimensional covariance matrices in the GMMs).

A first simple step to reduce the computational cost is to use the minibatching \cite{hoffman2013stochastic}, i.e.\ we compute the conditional expectation in \eqref{eq:intermediate_quantity} only for a subset---here referred to as a \emph{minibatch}---of the original full dataset, $(x_i)_{i\in I}$, at each $t$. Here, $I$ is a set of $B\ll N$ indices, $i$, sampled uniformly without replacement from $(1,\ldots,N)$. Though minibatching is well-known in the machine learning community, it was applied only recently in the context of the MCMCSAEM algorithm \cite{kuhn2020properties}.

We then continue to decrease the computational cost by sampling $M\ll K$ random samples from $p_\theta(z_i|x_i)$, for each $i\in I$, to obtain the Monte Carlo average, $\bar{\mathcal{Q}}_{t}$, in \eqref{eq:stochastic_approximation_generic}. Note that direct sampling from $p_\theta(z_i|x_i)$ would not lead to any substantial decrease in the number of operations, since we have to first compute the normalizing factor, $p_\theta(x_i)$. This requires $K$ expensive evaluations of $p_\theta(z_i,x_i)$, which is precisely what we want to avoid. The MCMC sampling in \eqref{eq:s_step_generic} allows us to sample from $p_\theta(z_i|x_i)$, with the cost decreasing to only $M\ll K$ evaluations of $p_\theta(z_i,x_i)$ per iteration.

A specific form of $P_\theta$ in \eqref{eq:s_step_generic} determines a resulting MCMC procedure. We chose the Metropolis-Hastings (MH) sampler (hence MHSAEM) which represents $P_{\theta_t}(z^{j-1}_{i,t}, z^j_{i,t})$ as follows: given $\bar{z}\coloneqq z^{j-1}_{i,t}$, draw a sample from the \emph{proposal} distribution, $z\sim q(\cdot|\bar{z})$, compute the acceptance ratio,
\begin{equation}
    \alpha(\bar{z},z)\coloneqq\operatorname{min}\!\bigg\lbrace 1,\frac{p_{\eta_{z,t}}(x_i|z)\pi_{z,t}q(\bar{z}|z)}{p_{\eta_{\bar{z},t}}(x_i|\bar{z})\pi_{\bar{z},t}q(z|\bar{z})}\bigg\rbrace
    ,
    \label{eq:acceptance_ratio}
\end{equation}
and, if $u<\alpha(\bar{z},z)$---where $u$ is drawn from the uniform distribution, $\text{Uniform}(0,1)$---accept the sample and set $z^j_{i,t}=z$; otherwise, set $z^j_{i,t}=\bar{z}$. We repeat this process for each $j\in(1,\ldots,M)$, construing a set $\mathbf{z}_{i,t}=(z^1_{i,t},\ldots,z^M_{i,t})$. Recall that we set $z^0_{i,t}\coloneqq z^M_{i,t-1}$ at each iteration, i.e.\ the chain has the length $MT$ (Section \ref{sec:mcmcsaem}). The samples, $\mathbf{z}_{i,t}$, are then used to obtain the following MC average:
\begin{equation}
    \bar{\mathcal{Q}}_{t}(\theta)=\frac{1}{M}\sum_{i\in I}\sum_{z\in \mathbf{z}_{i,t}}\log p_{\eta_z}(x_i|z)\pi_{z}
    .
    \label{eq:intermediate_quantity_specific}
\end{equation}
Using \eqref{eq:intermediate_quantity_specific} to directly approximate \eqref{eq:generic_e_step} is inefficient and impractical (Section \ref{sec:mcmcsaem}). Therefore, we utilize \eqref{eq:intermediate_quantity_specific} in a type of stochastic approximation, as detailed in Section \ref{sec:m_step}.

\subsection{M-step}\label{sec:m_step}
If the M-step \eqref{eq:m_step} cannot be computed under a closed-form solution, one can resort to direct gradient-descent optimization of $\mathcal{Q}(\theta)$, where $\operatorname{arg\hspace{1pt}max}$ is replaced by one (or more) step(s) of a gradient-descent technique. The EM algorithm is then referred to as the generalized EM algorithm \cite{wu1983convergence}. To the best of our knowledge, this extension has not yet been applied in the MCMCSAEM framework.

We replace \eqref{eq:stochastic_approximation_generic} by the stochastic gradient-descent method, $\theta_t=\theta_{t-1}+\gamma_t\nabla_\theta\bar{\mathcal{Q}}_{t}(\theta)$, where $\nabla_\theta$ is the gradient w.r.t.\ $\theta$. This is also a form of the stochastic approximation \cite{robbins1951stochastic}, where the computations made in $\nabla_\theta\bar{\mathcal{Q}}$ are accumulated via $\theta_t$ and reused over the iterations.

The parameters $\eta_z$ have a different form based on a specific case of $p_{\eta_z}(x|z)$, whereas the parameters $\pi_z$ of $p_{\pi_z}(z)$ form a fixed structure in \eqref{eq:incomplete_data_likelihood_generic}. Therefore, without loss of generality, we split \eqref{eq:m_step_generic} into a generic part and a fixed part,
\begin{subequations}
    \begin{align}
        \eta_{k,t}=\eta_{k,t-1}+\gamma_t\nabla_{\eta_k}\bar{\mathcal{Q}}_{t}(\theta)
        ,
        \\
        \nu_{k,t}=\nu_{k,t-1}+\gamma_t\nabla_{\nu_k}\bar{\mathcal{Q}}_{t}(\theta)
        ,
    \end{align}
    \label{eq:stochastic_approximation_specific}
\end{subequations}
\hspace{-4pt}where---to ensure that the probabilities, $(\pi_{k,t})^K_{k=1}$, satisfy the constraints (Section \ref{sec:problem_formulation})---we transform $\nabla_{\pi_{k}}\bar{\mathcal{Q}}$ via $\nu_{k}=\log\pi_{k}$ and optimize w.r.t.\ $\nu_{k}$. Then, to obtain $(\pi_{k,t})^K_{k=1}$ from $\bm{\nu}_t\coloneqq(\nu_{k,t})^K_{k=1}$, we use the softmax function, i.e.\ $\pi_{k,t}\coloneqq\operatorname{softmax}(\bm{\nu}_t)_k\coloneqq\exp(\nu_{k,t})/\sum^K_{l=1}\exp(\nu_{l,t})$.

The M-step \eqref{eq:m_step} is also computationally costly for large~$K$. This holds even when it can be reduced to closed-form updates of expected sufficient statistics with $p_{\eta_z}(x|z)$ belonging to the exponential family \cite{nguyen2020mini} (again, due to high $D$). Note that computing the gradients for all pairs in $(\nu_k,\eta_k)^K_{k=1}$ would be inefficient, especially since $\mathbf{z}_{i,t}$ contains only a small number of unique values of $\mathsf{Z}$ for $M\ll K$. Therefore, we further reduce the computational cost by computing $\nabla_{\eta_k}\bar{\mathcal{Q}}$ and $\nabla_{\nu_k}\bar{\mathcal{Q}}$ only for $k\in\text{unique}(\mathbf{z}_{i,t})$. By this last step, we achieved the sought decrease in the complexity of the EM algorithm from $\mathcal{O}(TDNK)$ to $\mathcal{O}(TDBM)$, where $B\ll N$ and $M\ll K$. We summarize the proposed approach in Algorithm \ref{alg:mhsaem}.

\begin{algorithm}[tb]
    \caption{The generalized MHSAEM algorithm}
    \label{alg:mhsaem}
    \textbf{Input}: $\theta_1$, $(z^0_{i,1})^N_{i=1}$, $(x_i)^N_{i=1}$ \hfill\textbf{Output}: $(\theta_t)^T_{t=1}$\par
    \begin{algorithmic}
        \FOR{$t\in(1,\ldots,T)$ or until convergence}
            \STATE form the set $I=(i_j)^B_{j=1}$ by sampling $i_j\sim(1,\ldots,N)$
            \FOR{$i\in I$}
                \STATE set $z^0_{i,t}\coloneqq z^M_{i,t-1}$
                \FOR{$j\in(1,\ldots,M)$}
                    \STATE set $\bar{z}\coloneqq z^{j-1}_{i,t}$
                    \STATE sample $z\sim q(z|\bar{z})$
                    \STATE sample $u\sim\text{Uniform}(0,1)$
                    \STATE compute $\alpha(\bar{z},z)$ in \eqref{eq:acceptance_ratio}
                    \IF{$u<\alpha(\bar{z},z)$}
                        \STATE set $z^j_{i,t}\coloneqq z$ and $\bar{z}\coloneqq z$
                    \ELSE
                        \STATE set $z^j_{i,t}\coloneqq\bar{z}$
                    \ENDIF
                \ENDFOR
                \STATE set $\mathbf{z}_{i,t}\coloneqq(z^1_{i,t},\ldots,z^M_{i,t})$
            \ENDFOR
            \STATE compute \eqref{eq:intermediate_quantity_specific}
            \STATE compute \eqref{eq:stochastic_approximation_specific} for $k\in\text{unique}(\mathbf{z}_{i,t})$
            \STATE compute $\pi_{k,t}\coloneqq\operatorname{softmax}(\bm{\nu}_t)_k$ for $k\in\mathsf{Z}$
        \ENDFOR
    \end{algorithmic}
\end{algorithm}

\subsection{Proposal distribution}\label{sec:proposal_distribtion}
The proposal distribution impacts the convergence speed and computational cost of the proposed algorithm. Here, we discuss various possible choices of~$q(z|\bar{z})$.

\emph{Optimal proposal (\textbf{O}).} The optimal proposal distribution is $q(z|\bar{z})\coloneqq q(z)\coloneqq p_\theta(z|x)$. This ensures that the acceptance ratio \eqref{eq:acceptance_ratio} is $\alpha(\bar{z},z)=1$. However, before sampling from $p_\theta(z|x)$, there is the need to perform $K$ expensive evaluations of $p_\theta(z,x)$ to evaluate the normalizing factor of $p_\theta(z|x)$, which is the reason we resorted to the MH sampler in the first place. We consider this case only to set the upper limit on admissible computational cost and to study the impact of sub-optimal proposal distributions.

\emph{Uniform proposal (\textbf{U}).} The uniform distribution on the interval from $1$ to $K$, $q(z|\bar{z})\coloneqq q(z)\coloneqq \text{Uniform}(1,K)$, is the simplest and computationally cheapest variant. However, for high $K$, the MH kernel with this proposal will produce a poorly mixing Markov chain (Section 6.9.4 in \citet{robert2013monte}), since new samples are less likely to be accepted in this uniform case, causing slow convergence.

\emph{Tabular proposal with forgetting (\textbf{TF}).} The key requirement to design a proposal distribution is to restrict its computational cost between that of the U and O proposals. We satisfy this constraint by designing an independent proposal, $q(z|\bar{z})\coloneqq q_{\bm{\alpha}}(z)\coloneqq\mathcal{C}(\bm{\alpha})$, for each $i\in(1,\ldots,N)$, where $\mathcal{C}(\bm{\alpha}_i)\propto\Pi^K_{k=1}\alpha^{\mathds{1}(z_i=k)}_{k,i}$ is the categorical distribution with the weights $\bm{\alpha}_i\coloneqq (\alpha_{1,i},\ldots,\alpha_{K,i})$. We estimate them as $\bm{\alpha}_{i,t}\coloneqq\frac{n_{i,t}}{\sum^K_{k=1}n_{k,i,t}}$, $n_{i,t}=(\bm{1}-\mathbf{e}_{z_{i,t}}\gamma_t)\odot n_{i,t-1}+\gamma_t\mathbf{e}_{z_{i,t}}$, where $n_{k,i,t}$ is the $k$th entry of $n_{i,t}$, $\bm{1}$ is the vector of ones, $\mathbf{e}_{k}$ is the basis (one-hot) vector with one at $k$th position and zeros otherwise, and $\odot$ is the Hadamard product. We refer to this case as ``table with forgetting'' (TF), since it holds an $N\times K$ table in the memory and $\gamma_t$ is a forgetting factor. The full derivation is in the supplement.

\subsection{Convergence}
The key difference between \cite{kuhn2004coupling,kuhn2020properties} and our approach is that we perform \eqref{eq:m_step_generic} via direct, gradient-based optimization \eqref{eq:stochastic_approximation_specific}, rather than a restrictive closed-form solution. However, this implies that the convergence results presented in these papers do not hold in our case, and there are currently no results covering our extension in the literature. The closest framework---to analyse the convergence of stochastic approximations perturbed by Markov-dependent noise---is presented in \cite{andrieu2005stability}.

\begin{figure}
    \centering
    \hspace{-7pt}\input{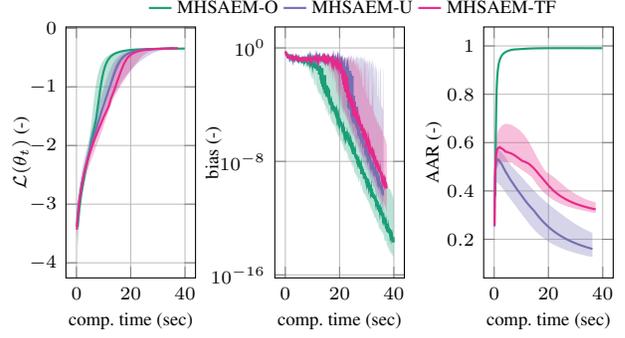}
    \caption{The training log-likelihood, $\mathcal{L}(\theta_t)$, (left) the bias, \eqref{eq:bias}, (middle) and the average acceptance ratio (right) versus the computational time (in seconds). Here, on the x-axis, the computational time at a current iteration, $t$, is obtained by accumulating the time from the previous iterations. The experiment settings are $(D,K,N,\omega,$ $B, M,T)$ $=(2,10,1k,$ $0.5,100,1,4k)$, see Section \ref{sec:experiments_gaussian} for details. The results are averaged over 50 repetitions with different initial conditions. The solid line is the median and the shaded area is the range from the 1st to 99th percentile.}
    \label{fig:convergence}
\end{figure}

Indeed, the gradients in \eqref{eq:stochastic_approximation_specific} are estimated via a non-homogeneous Markov chain (Section \ref{sec:mcmcsaem}), making the convergence analysis difficult. Therefore, we analyse the convergence empirically by computing the bias of the gradient estimator during the training. We use the fact that \eqref{eq:intermediate_quantity_specific} is the approximation of the exact EM objective (its minibatch version) which computes the expectation for each $\bar{z}\in\mathsf{Z}$,
\begin{align}
    ||\nabla_\theta\bar{\mathcal{Q}}_t-\nabla_\theta\bar{\mathcal{Q}}^{\text{exact}}_t||^2
    &\hspace{-1pt}=\hspace{-1pt}
    \textstyle
    ||\frac{1}{M}\sum_{i\in I}(\sum_{z\in\mathbf{z}_{i,t}}\hspace{-6pt}\nabla_\theta\log p_{\theta}(x_i,z)
    \nonumber
    \\
    &\hspace{-50pt}
    \textstyle
    -\sum_{\bar{z}\in\mathsf{Z}}\nabla_\theta\log p_{\theta}(x_i,\bar{z})p_{\theta_t}(\bar{z}|x_i))||^2
    ,
    \label{eq:bias}
\end{align}
where $||\cdot||$ is the $l_2$-norm. We perform a toy experiment with the settings from Section \ref{sec:experiments_gaussian} (except that we do not use the sufficient statistics). To avoid the impression that the method converges due to the decreasing step-size, we set $\gamma_t=0.01$ for $t\in(1,\ldots,T)$. Figure \ref{fig:convergence} shows that, even for $M=1$, the bias converges to zero as the iterations increase, and the training log-likelihood reaches a local optimum, which is the case for all proposals from Section \ref{sec:proposal_distribtion}. This empirically confirms that the E-step is asymptotically unbiased and suggests the validity of the algorithm. We also show the average acceptance ratio (AAR) over all $i\in(1,\ldots,N)$ at each iteration, $t$, i.e.\ we use \eqref{eq:acceptance_ratio} to compute $\text{AAR}_t\coloneqq\frac{1}{N}\sum^N_{i=1}\alpha(z_{i,t-1},z_{i,t})$.

\section{Related work}\label{sec:related_work}
\emph{Stochastic approximation expectation-maximization.} The application of SA to prevent the evaluation of all $K$ components in mixture models has been overlooked for a long time. The reason is that the main motivation to combine the EM algorithm with SA is to address the analytical intractability of the expected value under $p_\theta(z|x)$ in \eqref{eq:intermediate_quantity}, which is, however, always tractable for mixture models. The intractability issue is addressed by either the Monte Carlo SAEM (MCSAEM) \cite{delyon1999convergence} or the Markov chain Monte Carlo SAEM (MCMCSAEM) \cite{kuhn2004coupling}. Applying the former approach to mixture models is inefficient, since it evaluates $K$ joint distributions, $p_\theta(z,x)$, before drawing $M$ samples from $p_\theta(z|x)$. Therefore, this method reduces only the computational cost of updating the \emph{sufficient statistics}. This is addressed by the latter approach, where $M<K$ samples from a proposal distribution, $q(z|x)$, is used to calculate $p_\theta(z,x)$ and also the sufficient statistics. An alternative way to achieve this would be to use the importance sampling \cite{levine2001implementations}. However, all these methods process all data at every iteration, providing a limited advantage over the EM algorithm. Minibatch versions of these techniques have recently been proposed \cite{karimi2019convergence,kuhn2020properties,allassonniere2021new}.

All the above methods assume $p_\theta(z,x)$ belonging to the exponential family. This is a convenient, but limiting, property which allows \eqref{eq:m_step} to be computed under a closed-form solution. The main contribution of our work is to release this restrictive assumption by admitting that $p_\theta(z,x)$ (and thus $\mathcal{Q}$) is given by complex and intractable transformations.

\emph{Sparse and truncated variational techniques.} There is only a few methods specifically tailored to evaluate less components. They commonly follow from the variational framework, where the exact posterior, $p_\theta(z|x)$, is approximated by a variational one, $q(z|x)$. This approximation is defined over $M\ll K$ components such that only the important ones are selected, relying on relaxation of the hard EM algorithm from taking a single $M=1$ \cite{juang1990segmental} to multiple $M\ll K$ components. The sparse SAEM (SSAEM) algorithm \cite{hughes2016fast} selects the components by a quick partial sorting of the posterior probabilities, $p_\theta(z|x)$. Again, this requires $K$ evaluations of $p_\theta(z,x)$ before the sorting, thus only reducing the amount of updated statistics. The truncated SAEM (TSAEM) algorithm \cite{forster2018can} selects $M<K$ cluster-to-cluster and $\bar{M}<K$ cluster-to-datapoint minimal distances, preventing the problem in the SSAEM algorithm. However, the distances are evaluated for all combinations of components, leading to $K^2$-computational cost, which makes the saving dubious. Moreover, these methods also assume that $p_\theta(z,x)$ is from the exponential family.

We summarize the distinguishing features of the above discussed methods in Table \ref{tab:algorithms}. We defer the alternative methods, which decrease the computational cost via the first three factors in $\mathcal{O}(TDNK)$, into the supplement (Section 12).

\section{Experiments}\label{sec:experiments}
To demonstrate the key features of our algorithm---its low computational cost, competitive learning performance, and generality---we use it below to train: (i) GMMs on synthetic data, and (ii) SPTNs \cite{pevny2020sum} and (iii) mixtures of real NVP flows \cite{pires2020variational} on real data. 
All experiments have been performed on a Slurm cluster equipped with Intel Xeon Scalable Gold 6146 with 384GB of RAM. Additional details on experiment settings are presented in Section 11 of the supplement.

\begin{figure}
    \centering
    \input{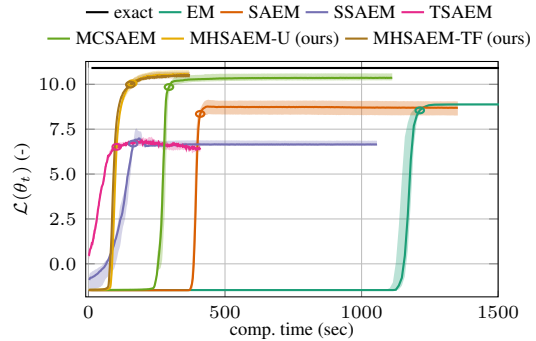}
    \caption{The training log-likelihood, $\mathcal{L}(\theta_t)$, versus the computational time (in seconds). Here, on the x-axis, the computational time at a current iteration, $t$, is obtained by accumulating the time from the previous iterations. \ref{circle} corresponds to $\mathcal{L}(\theta_{t_{95}})$, where $t_{95}$ is the iteration of reaching 95\% of $\operatorname{max}\mathcal{L}(\theta_{t})$. The projection of \ref{circle} on the x-axis gives the time to reach $\mathcal{L}(\theta_{t_{95}})$. The experiment settings are $(D,K,N,\omega,$ $B, M,T)$ $=(10,100,10k,$ $0.1,200,2,20k)$, see Section \ref{sec:experiments_gaussian} for details. The results are averaged over five repetitions with different initial conditions. The solid line is the median and the shaded area is the range from the 1st to 99th percentile.}
    \label{fig:loglik_vs_computational_time}
\end{figure}

\subsection{Gaussian mixture models}\label{sec:experiments_gaussian}
Let us consider that the components in \eqref{eq:incomplete_data_likelihood_generic} take the form of the Gaussian distribution, $p_{\eta_z}(x|z)=\mathcal{N}(x;\mu_z, \Sigma_z)$, where $\mu_z$ is the mean vector and $\Sigma_z$ is the covariance matrix.

\begin{figure}
    \centering
    \hspace{-5pt}\input{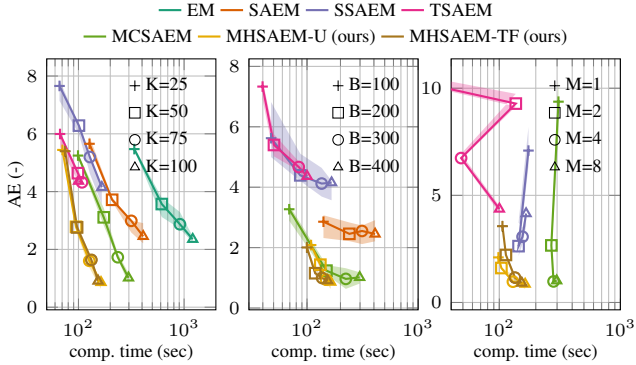}
    \caption{The absolute error, $\text{AE}=|\mathcal{L}(\theta_{t_{95}})-\mathcal{L}(\theta)|$, versus the computational time (in seconds). The experiment settings are $(D,K,N,\omega,$ $B, M,T)$ $=(10,100,10k,$ $0.1,200,2,20k)$, where the number of components, $K$, (left), the batchsize, $B$, (middle) and the number of samples, $M$, (right) change for different values denoted by (\ref{cross}, \ref{square}, \ref{circle}, \ref{triangle}). At each of these points, we perform an experiment as illustrated in Figure \ref{fig:loglik_vs_computational_time}, find $\mathcal{L}(\theta_{t_{95}})$ to compute the AE, and record the time of reaching $t_{95}$. The solid line is the median and the shaded area is the range from the 1st to 99th percentile over five repetitions with different initial conditions.}
    \label{fig:mae_vs_computational_time}
\end{figure}

\textit{Experiment settings:} We generate the parameters of \eqref{eq:incomplete_data_likelihood_generic}, and the corresponding dataset, uniquely for a given quadruple $(D,K,N,\omega)$, where $\omega$ is the maximum pairwise overlap between the components (see \citet{melnykov2012mixsim} and the supplement for details). Since the parameters of the generative model are known, we can track the convergence of the training log-likelihood, $\mathcal{L}(\theta_t)$, compared to the exact one, $\mathcal{L}(\theta)$, for $t=(1,\ldots,T)$. We are interested in the absolute error between the training log-likelihood at the iteration of reaching 95\% of its maximum value, $t_{95}$, and the exact log-likelihood, i.e.\ $\text{AE}=|\mathcal{L}(\theta_{t_{95}})-\mathcal{L}(\theta)|$. We also measure the computational time until reaching $t_{95}$. We use 95\% of the maximum value instead of the maximum value to prevent the oscillations around a stationary point, making the convergence time very noisy (e.g.\ SAEM in Figure~\ref{fig:loglik_vs_computational_time}).

\textit{Algorithms:} The joint likelihood, $p_\theta(z,x)$, of the GMM belongs to the exponential family. Therefore, we use a closed-form solution of \eqref{eq:stochastic_approximation_generic}, \cite{nguyen2020mini}, and the convergence results from \cite{kuhn2020properties} apply in this case. We compare our MHSAEM algorithm with the variational algorithms in Table \ref{tab:algorithms}. To counteract poor local optima, we apply the deterministic anti-annealing to all algorithms in Table \ref{tab:algorithms}, as detailed in \cite{naim2012convergence} and the supplement (which contains the result without the annealing). Note that we do not use the O-proposal (Section \ref{sec:proposal_distribtion}), since the MHSAEM-O algorithm is equivalent to the MCSAEM algorithm.

\textit{Results:} Figure \ref{fig:loglik_vs_computational_time} shows that the EM and SAEM algorithms attain a poor local optima in the longest time. In contrast, the MCSAEM and MHSAEM (U and TF) algorithms achieve $\mathcal{L}(\theta_{t_{95}})$ that is closest to $\mathcal{L}(\theta)$ of the true model. The reason is that these methods have the additional stochastic source---i.e.\ the sampling of $z$ (in addition to minibatching)---which improves the exploration of the likelihood surface. Note, too, that the MHSAEM algorithms get closest to $\mathcal{L}(\theta)$ in the shortest time compared to all the other methods. The SSAEM and TSAEM algorithms provide a comparable computational time, but they have the lowest $\mathcal{L}(\theta_{t_{95}})$. This is caused by selecting only $M$ maximal probabilities in the SSAEM (or distances in the TSAEM) algorithm (Section \ref{sec:related_work}), which prevents certain, but not a negligible number of, components from being updated, providing only a crude approximation of $p_{\theta}(z|x)$. The algorithms then suffer from a substantial variational gap w.r.t.\ $\mathcal{L}(\theta)$.

In Figure \ref{fig:mae_vs_computational_time}, we investigate sensitivity of fitting the model to increasing values of $K$, $B$ and $M$. In all the cases, our algorithms achieve the lowest AE in the shortest time, whereas the SSAEM and TSAEM algorithms fail to converge for $M>2$ and $K>50$, respectively. This behaviour follows from the same explanation as in the previous paragraph.

\subsection{Sum-product transform networks}\label{sec:experiments_sptn}
The sum product networks (SPNs) are a deep learning extension of finite mixture models. They can be interpreted as a mixture of trees \cite{zhao2016unified}, where each tree corresponds to a component. Therefore, they can be cast into the form of \eqref{eq:incomplete_data_likelihood_generic}, but the number of components grows exponentially with their depth. In this section, we use recently proposed SPTNs which introduce additional transformation nodes to provide better expressiveness than SPNs.

\setlength{\tabcolsep}{4pt}
\begin{table*}
  \centering
  \small
  \begin{tabular}{r|rrrrr|rrrrr}
    \hline\hline
                            & \multicolumn{5}{c|}{Sum-product transform networks}                                        & \multicolumn{5}{c}{Mixtures of real NVP flows}\\\cline{2-11}
                            &                            & \multicolumn{2}{c}{\textbf{SGD}}      & \multicolumn{2}{c|}{\textbf{MHSAEM-U}} & & \multicolumn{2}{c}{\textbf{SGD}} & \multicolumn{2}{c}{\textbf{MHSAEM-U}}\\
    \textbf{dataset}        & \textbf{speed-up}          &  $\mathcal{L}^{\text{test}}$          & $K$  & $\mathcal{L}^{\text{test}}$             &  $K$  & \textbf{speed-up}          & $\mathcal{L}^{\text{test}}$            & $K$ & $\mathcal{L}^{\text{test}}$             & $K$\\\hline
    breast-cancer           & 4.66                       & -4.66$\pm$3.07                        & 64   & \color{blue}{\textbf{1.43$\pm$1.70}}    & 1024  & \color{red}{\textbf{0.63}} & -99.85$\pm$38.45                       & 32  & \color{blue}{\textbf{-39.31$\pm$1.07}}  & 128\\
    cardio                  & 10.55                      & \color{blue}{\textbf{59.52$\pm$3.83}} & 512  & 31.04$\pm$4.23                          & 1024  & 9.85                       & 54.34$\pm$6.48                         & 32  & \color{blue}{\textbf{56.08$\pm$4.56}}   & 128\\
    telescope               & 102.53                     & \color{blue}{\textbf{-3.65$\pm$0.08}} & 512  & -5.03$\pm$0.06                          & 1024  & 3.74                       & \color{blue}{\textbf{-3.97$\pm$0.12}}  & 8   & -4.22$\pm$0.07                          & 8  \\
    pendigits               & 4.89                       & \color{blue}{\textbf{0.88$\pm$0.07}}  & 1024 & -4.86$\pm$0.46                          & 16384 & 4.17                       & \color{blue}{\textbf{1.46$\pm$0.59}}   & 8   & 0.48$\pm$0.56                           & 8  \\
    pima-indians            & \color{red}{\textbf{0.37}} & -8.54$\pm$0.39                        & 64   & \color{blue}{\textbf{-7.62$\pm$0.81}}   & 64    & 1.35                       & -20.09$\pm$13.13                       & 128 & \color{blue}{\textbf{-16.33$\pm$0.39}}  & 128\\
    robot                   & 3.43                       & \color{blue}{\textbf{1.84$\pm$3.18}}  & 1024 & -11.3$\pm$0.40                          & 16384 & 22.21                      & \color{blue}{\textbf{-14.26$\pm$1.29}} & 128 & -17.56$\pm$1.20                         & 128\\
    waveform-1              & 4.35                       & -26.14$\pm$0.39                       & 64   & \color{blue}{\textbf{-23.91$\pm$0.07}}  & 1024  & 3.72                       & -34.12$\pm$0.98                        & 8   & \color{blue}{\textbf{-33.42$\pm$0.84}}  & 8  \\
    waveform-2              & 4.82                       & -26.21$\pm$0.43                       & 64   & \color{blue}{\textbf{-23.91$\pm$0.05}}  & 1024  & 4.12                       & -34.15$\pm$0.80                        & 8   & \color{blue}{\textbf{-33.64$\pm$0.65}}  & 8  \\
    yeast                   & 20.57                      & \color{blue}{\textbf{10.26$\pm$0.65}} & 512  & 5.18$\pm$0.46                           & 1024  & 14.49                      & 6.61$\pm$4.11                          & 128 & \color{blue}{\textbf{9.59$\pm$3.75}}    & 128\\
    ecoli                   & 1.86                       & -5.50$\pm$1.52                        & 64   & \color{blue}{\textbf{-0.22$\pm$0.65}}   & 1024  & 2.15                       & -11.37$\pm$8.82                        & 128 & \color{blue}{\textbf{-10.64$\pm$4.64}}  & 128\\
    ionosphere              & 1.88                       & -20.27$\pm$5.61                       & 64   & \color{blue}{\textbf{-5.93$\pm$2.19}}   & 512   & 2.74                       & -87.01$\pm$31.67                       & 128 & \color{blue}{\textbf{-42.75$\pm$2.28}}  & 128\\
    iris                    & \color{red}{\textbf{0.23}} & -10.65$\pm$0.95                       & 64   & \color{blue}{\textbf{-1.49$\pm$0.00}}   & 16384 & 3.28                       & -16.34$\pm$10.13                       & 128 & \color{blue}{\textbf{-9.21$\pm$1.23}}   & 32 \\
    blocks                  & 12.18                      & \color{blue}{\textbf{12.21$\pm$1.03}} & 512  & 6.84$\pm$0.42                           & 1024  & 44.95                      & 17.13$\pm$0.37                         & 128 & \color{blue}{\textbf{17.94$\pm$0.56}}   & 32 \\
    parkinsons              & 1.46                       & -21.85$\pm$3.51                       & 64   & \color{blue}{\textbf{0.5$\pm$1.47}}     & 512   & 3.09                       & -566.58$\pm$379.52                     & 128 & \color{blue}{\textbf{-33.31$\pm$1.66}}  & 32 \\
    sonar                   & 2.96                       & -95.39$\pm$4.42                       & 512  & \color{blue}{\textbf{-69.29$\pm$10.03}} & 64    & 2.52                       & -622.2$\pm$630.50                      & 128 & \color{blue}{\textbf{-88.81$\pm$4.22}}  & 128\\
    segment                 & 1.44                       & \color{blue}{\textbf{47.35$\pm$3.89}} & 512  & 26.53$\pm$2.30                          & 16384 & 38.49                      & 35.84$\pm$3.98                         & 128 & \color{blue}{\textbf{42.04$\pm$3.74}}   & 32 \\
    vehicle                 & 2.97                       & \color{blue}{\textbf{-4.25$\pm$0.86}} & 64   & -5.45$\pm$0.34                          & 1024  & 6.78                       & -31.34$\pm$2.06                        & 32  & \color{blue}{\textbf{-26.43$\pm$0.77}}  & 128\\
    wine                    & 75.42                      & -25.99$\pm$0.44                       & 1024 & \color{blue}{\textbf{-13.27$\pm$0.54}}  & 1024  & 2.05                       & -171.58$\pm$60.12                      & 128 & \color{blue}{\textbf{-25.57$\pm$0.54}}  & 128\\\hline
    rank                    &                            & 1.56                                  &      & \color{blue}{\textbf{1.44}}             &       &                            & 1.83                                   &     & \color{blue}{\textbf{1.17}}             &    \\\hline\hline
  \end{tabular}
  \caption{The speed-up and test log-likelihood, $\mathcal{L}^{\text{test}}$, for the SGD and MHSAEM-U algorithms. The test log-likelihood (higher is better) is computed for the best model, with the corresponding $K$, which is selected based on the validation log-likelihood. The speed-up is computed as the ratio of MHSAEM-U to SGD, i.e.\ their time to reach 95\% of the training log-likelihood. The results are averaged over five repetitions with different initial conditions. The likelihood is shown with its standard deviation. Then, the higher test log-likelihood is highlighted with bold blue, and and no speed-up is highlighted with red. The average rank is computed as the standard competition (``1224'') ranking \cite{demvsar2006statistical} on each dataset (lower is better).}
  \label{tab:datasets_sptn}
\end{table*}

\textit{Experiment settings:} We use 19 real datasets from the UCI database \cite{Dua2019UCI,mangasarian1990cancer,little2007exploiting,siebert1987vehicle}, preprocessed in the same way as in \cite{pevny2016loda}. For each experiment, we randomly split the data into 64\%, 16\% and 20\% for training, validation and testing, respectively. We calculate the average log-likelihood on the test set and measure again the time to reach 95\% of the maximal training log-likelihood, $\mathcal{L}(\theta_{t_{95}}).$

To evaluate various (shallow and/or deep) architectures of SPTNs, we fit each dataset with the following combinations of hyper-parameters\footnote{We have set a hard limit to train a single model to 24h, which is default on our Slurm cluster.}: $s\in(8,32,128)$, $b\in(2,4,6,8)$, $l\in(2,3,4)$, where $s$ is the number of children of each sum node, $b$ is the number of partitions of each product node, and $l$ is the number of layers (one layer contains sum and product nodes). The number of components of the SPTN, after its conversion into \eqref{eq:incomplete_data_likelihood_generic}, is $K=s^l$.

\textit{Algorithms:} The TF-proposal does not deliver much better performance than the U-proposal for high $K$ (Figure \ref{fig:mae_vs_computational_time}). Moreover, the U-proposal is easier to implement. Therefore, we use only the MHSAEM-U algorithm and compare it to the stochastic gradient-descent (SGD) algorithm, which is routinely applied to training of SP(T)Ns \cite{peharz2020random,pevny2020sum}. The SGD algorithm performs computations over all subtrees of the network, whereas the MHSAEM-U algorithm uses only $M=1$ subtree, thus we should observe a speed-up of the computations.

\textit{Results:} Since each dataset might benefit from a different architecture, Table \ref{tab:datasets_sptn} (left) shows the test log-likelihood of the architectures selected via the best log-likelihood measured on the validation set and the associated speed-up. The test log-likelihoods reveal that the MHSAEM-U algorithm outperforms the SGD algorithm on 10 out of 19 datasets, which was not originally the goal. Similarly to Section \ref{sec:experiments_gaussian}, the success lies behind the additional stochastic source which helps to escape poor local optima. This also explains why there are sometimes big differences between the log-likelihoods of the MHSAEM-U and SGD algorithms. The speed-up demonstrates lower computational cost of the MHSAEM-U algorithm on 17 out of 19 datasets, which was the main goal. The \texttt{telescope} and \texttt{wine} datasets show approximately 102$\times$ and 75$\times$ speed-up, respectively, while on very small datasets (\texttt{pima-indians} and \texttt{iris}), the SGD is faster due to effective implementation. We can also observe that the standard deviations of the MHSAEM-U algorithm are lower on most of the datasets. Table 3 in the supplement shows the same trends for a fixed architecture.

\subsection{Mixtures of real NVP flows}\label{sec:experiments_real_nvp}
We consider another class of mixture models~\eqref{eq:incomplete_data_likelihood_generic}, where each $p_{\eta_z}(x|z)$ is transformed by the flow model---real NVP \cite{dinh2017density}. The transformations are given by deep neural networks, allowing for flexible adjustment of the learning capacity of each component.

\textit{Experiment settings:} We use the same experiment settings and metrics from Section \ref{sec:experiments_sptn}, changing the number of components as follows: $K\in(8,32,128)$. We have used the implementation from~\cite{francu2020}.

\textit{Algorithms:} We adopt the same methods as in Section~\ref{sec:experiments_sptn}.

\textit{Results:} The results are presented in Table \ref{tab:datasets_sptn} (right). They are similar to those obtained in the previous section. In terms of the test log-likelihood, the MHSAEM-U algorithm outperforms the SGD algorithm on all but three datasets, and it provides a substantial speed-up on all datasets except one. The test likelihood of models with the real NVP flows is most of the time worse than that of SPTNs with the affine transformations. As explained in the supplement, this is due to the overfitting, which was observed in~\cite{pevny2020sum}.

\section{Conclusion}
This paper has presented a method to decrease computational cost of fitting mixture models, including their generalizations, such as sum-product-(transform) networks and mixtures of flow models. The speed-up is achieved by evaluating only a single component (per iteration) via the MH sampling. The comparison on all three aforementioned models confirmed that the method significantly speeds-up the fitting time, and, importantly, without sacrificing the quality of the fit. In fact, the likelihood was better than that of the models fitted by the EM algorithm or the SGD algorithm in more than 50\% of cases. We attribute this to higher stochasticity, which helps to escape from poor local optima.

In the experiments, the proposed method has used a uniform proposal distribution. Despite outperforming the alternative methods, we conjecture that this limits the speed of convergence. Therefore, we believe that there is still a room for improvement in the implementation. We will address this in future work. We see the main limitation of our method in its extension into the online regime, where $x_i$ arrives sequentially. The reason is that $z_i$ does not exhibit Markov dependence under $i$ for the considered models.


\balance
\small
\bibliography{aaai22}

\newpage
\normalsize

\onecolumn
\begin{center}
	\textbf{\LARGE Supplementary Material - Fitting Large Mixture Models\\Using Stochastic Component Selection}
\end{center}
\vspace{10pt}

\noindent
This supplementary material provides additional details and experiments for the paper entitled: \emph{``Fitting large mixture models using stochastic component selection''}. The code to reproduce all experiments in the main paper and this supplement is available at: \url{https://drive.google.com/drive/folders/1foHlyCCHS8N2Odw6sXQI9bDCRTQIuwkU?usp=sharing}
\vspace{10pt}

\begin{multicols}{2}
\section{Gaussian mixture models}
In this section, we provide more experiments with the GMMs presented in Section 6.1.

\subsection{Counteracting weak local optima with deterministic annealing}
The EM algorithm converges to a local optimum in a finite number of iterations \citesupp{wu1983convergence}. In simple scenarios---where the number of components, $K$, is low---making a few attempts with different initial conditions may successfully lead to finding the global optimum. However, for large $K$, the likelihood surface is complicated and contains several weak local optima. To counteract this issue, we apply the deterministic annealing \citesupp{ueda1994deterministic}, which---similarly to the simulated annealing \citesupp{geman1984stochastic}---has its roots in thermodynamics and the maximum entropy principle. Indeed, maximizing the entropy term, $\mathcal{H}(\hat{\theta})$, in (3) of the main paper w.r.t.\ a variational posterior distribution, $q_{\hat{\theta}}(z|x)$, leads to $q_{\hat{\theta}}(z|x)\propto p_{\hat{\theta}}(z,x)^{\beta}$, where $\beta$ is the inverse temperature parameter. This does not change the form of the EM objective, $\mathcal{Q}(\theta)$, expect that its original posterior, $p_{\hat{\theta}}(z|x)$, is replaced by $q_{\hat{\theta}}(z|x)$. Then, it can be shown that $\beta$ modifies the log-likelihood, $\mathcal{L}(\theta)$, \citesupp{ueda1994deterministic}. Specifically, for $\beta\rightarrow 0$ (high temperature), the likelihood surface is nearly uniform, having a single global optimum, whereas for $\beta\rightarrow 1$ (low temperature), $q_{\hat{\theta}}(z|x)\rightarrow p_{\hat{\theta}}(z|x)$, having several local optima. Therefore, the key requirement is to change $\beta$ via a pre-specified annealing schedule, $(\beta_t)^T_{t=1}$, such that the global optimum slowly appears and is thus easier to find. This principle is applicable to all the algorithms in Table 1.

We resort to the deterministic anti-annealing \citesupp{naim2012convergence}, which improves the convergence speed over the deterministic annealing by admitting $\beta>1$. Specifically, it forms $(\beta_t)^T_{t=1}$ as follows: start with $\beta_{1}=\beta_{\text{min}}$, then increase to $\beta_{\tau}=\beta_{\text{max}}>1$, where $\tau<T$, and finally decrease back to $\beta_{T}=1$.

\begin{figure*}
	\centering
	\input{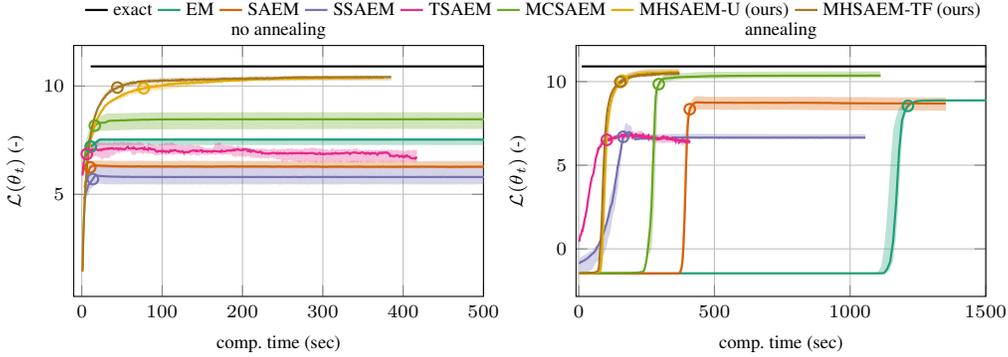}
	\caption{The training log-likelihood, $\mathcal{L}(\theta_t)$, versus the computational time (in seconds). The algorithms either do not use (left) or use (right) the anti-annealing schedule described in Section 6.1. Here, on the x-axis, the computational time at a current iteration, $t$, is obtained by accumulating the time from the previous iterations. \protect\markit{o} corresponds to $\mathcal{L}(\theta_{t_{95}})$, where $t_{95}$ is the iteration of reaching 95\% of $\operatorname{max}\mathcal{L}(\theta_{t})$. The projection of \protect\markit{o} on the x-axis gives the time to reach $\mathcal{L}(\theta_{t_{95}})$. This experiment was performed with the following settings: $(D,K,N,\omega,B, M,T)$ $=(10,100,10k,0.1,200,2,20k)$. The results are averaged over five repetitions with different initial conditions. The solid line is the median and the shaded area is the range from the 1st to 99th percentile.}
	\label{fig:loglik_vs_computational_time_annealing_comparison}
\end{figure*}

Figure \ref{fig:loglik_vs_computational_time_annealing_comparison} compares the various EM algorithms in Table 1 with (right) and without (left) the anti-annealing schedule. We observe that the EM, SAEM and MCSAEM algorithms with the annealing find a better local optimum compared to the corresponding counterparts without the annealing. However, the annealing seems to have no effect on the poor optimum attained by the SSAEM and TSAEM algorithms. Similarly, there is only a slight improvement in the optimum reached by the MHSAEM algorithm (both U and TF proposals). Still, the MHSAEM algorithm delivers the best performance compared to all the other algorithms, having the estimated log-likelihood very close to the exact one. We attribute the small distance to the exact log-likelihood to the overlap of a small portion of less representative clusters, preventing them from being represented via their sufficient statistics to an adequate degree.

Overall, albeit the annealing enhances the algorithms' ability to find a better optimum, it is rather the stochastic nature of the sampling-based algorithms what provides a better fit of the model.

\subsection{The computational cost under various operating conditions}
Recall that the computational cost of our MHSAEM algorithm is $\mathcal{O}(DTBM)$ (Section 4), where $D$ is the dimension of data, $T$ is the total amount of iterations, $B$ is the batchsize, and $M$ is the number of samples. Although there is no direct dependence on $K$, certain aspects of the algorithm still do depend on $K$, e.g.\ all operations associated with the memory management of selecting $M$ of $K$ sufficient statistics. Therefore, we investigate how the computational cost and the training performance depend on $K$ while changing $D$, $M$ and $B$. We leave $T$, since this affects all algorithms in the same way. We use the metrics from Section~6.1.

\begin{figure*}
	\centering
	\input{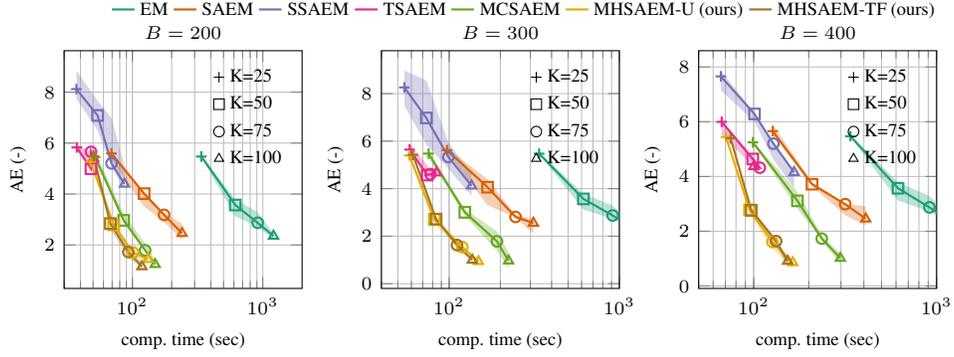}
	\vspace{-8pt}
	\caption{The absolute error, $\text{AE}=|\mathcal{L}(\theta_{t_{95}})-\mathcal{L}(\theta)|$, versus the computational time (in seconds). All experiments use the following settings: $(D,K,N,\omega,B, M,T)$ $=(2,K,20k,0.001,B,8,40k)$, where the number of components, $K$, changes for different values denoted by (\protect\markit{+}, \protect\markit{square}, \protect\markit{o}, \protect\markit{triangle}) and the batchsize is $B=200$ (left), $B=300$ (middle) and $B=400$ (right). At each of these points (marks), we perform an experiment as illustrated in Figure 1 (right), find $\mathcal{L}(\theta_{t_{95}})$ to compute the AE, and record the time corresponding to $t_{95}$. The results are averaged over five repetitions with different initial conditions. The solid line is the median and the shaded area is the range from the 1st to 99th percentile.}
	\label{fig:lkl_err_vs_ncomp_time_bsize}
\end{figure*}

\emph{Batchsize.} All the SA-algorithms in Table 1 sub-sample data with a minibatch of size $B$. The algorithms differ in a way they process the minibatch of datapoins, most notably in the number of evaluated components at each of the datapoints. Figure \ref{fig:lkl_err_vs_ncomp_time_bsize} shows the AE versus the computational time for various $K$ and $B$. Note we include the EM algorithm, which does not depend on $B$, just for a comparison. From all the $B$-dependent variants, the SAEM algorithm has the highest computational cost, since it processes all $K$ components for each datapoint in the minibatch. The MHSAEM algorithms attain the best AE in the shortest time, assessing only $M$ components for each datapoint in the minibatch. Although the MCSAEM algorithm provides a similar AE, its computational time grows faster with increasing $B$. This is due to that the posterior distribution has to be evaluated for all $K$ components before the MC sampling. The SSAEM algorithm preserves a similar computational cost as the MHSAEM algorithms for all $B$. However, it delivers a poor AE. The TSAEM algorithm fails to converge in this experiment.

\emph{Number of samples.} The number of samples (the MCSAEM and MHSAEM algorithms), or selection points (the SSAEM and TSAEM algorithms), $M$, determines the amount of components processed for each datapoint in the minibatch. In other words, this quantity determines the number of sufficient statistics to be updated at each iteration of a given algorithm. Since this feature is common across all $M$-dependent algorithms in Table 1, we can expect that the computational times will differ approximately by a constant factor, which is due to different sampling, or selection, mechanisms. Indeed, this is seen in Figure \ref{fig:lkl_err_vs_ncomp_time_nsamp}, where there is approximately the same distance between the MCSAEM and MHSAEM algorithms for all $M$. This can also be observed for the SSAEM algorithm, except the last case with $M=8$. We believe that this is caused by the application of the fast stochastic sorting algorithm \citesupp{hughes2016fast}, which increases the computational time due to a different initial state of the unsorted vector of the posterior probabilities at each iteration, $t$. Again, the TSAEM algorithm delivers a rather unsystematic behaviour. Similarly as before, we include the $M$-independent EM and SAEM algorithms for a comparison.

\emph{Dimension.} The dimension, $D$, has a direct impact on the computational cost of updating the expected sufficient statistics, evaluating the coditional likelihood, $p_{\hat{\theta}}(x|z)$, and computing the parameter estimates, $\hat{\theta}$. All the algorithms in Table 1 differ in the way they perform these elementary operations. Therefore, in Figure \ref{fig:lkl_err_vs_ncomp_time_ndims}, we compare the computational time for various choices of $D$. Unfortunately, the values, $D\in(2,4,8)$, are too low to demonstrate clear differences among the algorithms. This is caused by that the time to carry out the algebraic operations with such similar low dimensions is too short to stand out against the overhead associated with other computational steps in the algorithms. This is easy to deal with by repeating the experiment for, say, $D\in(1,10,100)$, or further optimizing our implementation. We will address this in future work. Nonetheless, this experiment shows consistent behaviour of our MHSAEM algorithm under varying $D$.

\begin{figure*}
	\centering
	\input{plots/lkl_err_vs_ncomp_time_nsamp.tikz}
	\vspace{-8pt}
	\caption{The absolute error, $\text{AE}=|\mathcal{L}(\theta_{t_{95}})-\mathcal{L}(\theta)|$, versus the computational time (in seconds). All experiments use the following settings: $(D,K,N,\omega,B, M,T)$ $=(2,K,20k,0.001,400,M,40k)$, where the number of components, $K$, changes for different values denoted by (\protect\markit{+}, \protect\markit{square}, \protect\markit{o}, \protect\markit{triangle}) and the number of samples is $M=2$ (left), $M=4$ (middle) and $M=8$ (right). At each of these points (marks), we perform an experiment as illustrated in Figure 1 (right), find $\mathcal{L}(\theta_{t_{95}})$ to compute the AE, and record the time corresponding to $t_{95}$. The results are averaged over five repetitions with different initial conditions. The solid line is the median and the shaded area is the range from the 1st to 99th percentile.}
	\label{fig:lkl_err_vs_ncomp_time_nsamp}
\end{figure*}

\begin{figure*}
	\centering
	\input{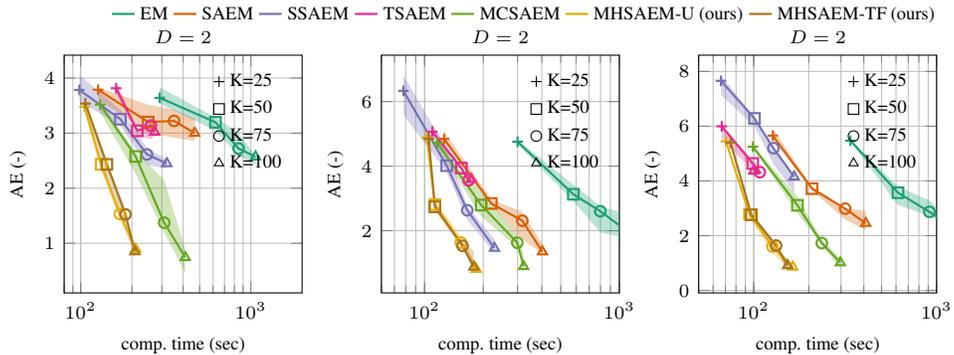}
	\vspace{-8pt}
	\caption{The absolute error, $\text{AE}=|\mathcal{L}(\theta_{t_{95}})-\mathcal{L}(\theta)|$, versus the computational time (in seconds). All experiments use the following settings: $(D,K,N,\omega,B, M,T)$ $=(D,K,20k,0.001,400,8,40k)$, where the number of components, $K$, changes for different values denoted by (\protect\markit{+}, \protect\markit{square}, \protect\markit{o}, \protect\markit{triangle}) and the dimension of data is $D=2$ (left), $D=4$ (middle) and $D=8$ (right). At each of these points (marks), we perform an experiment as illustrated in Figure 1 (right), find $\mathcal{L}(\theta_{t_{95}})$ to compute the AE, and record the time corresponding to $t_{95}$. The results are averaged over five repetitions with different initial conditions. The solid line is the median and the shaded area is the range from the 1st to 99th percentile.}
	\label{fig:lkl_err_vs_ncomp_time_ndims}
\end{figure*}

\subsection{Efficiency of the proposal distribution}
The proposal distribution is the key factor influencing the performance of the MH sampler. Therefore, we compare the proposal distributions discussed in Section 4.3. To demonstrate the impact of introducing the forgetting factor in the TF proposal, we define another tabular proposal, here abbreviated with T, which is given by $q_{\bm{\alpha}_i}(z_i)\coloneqq\mathcal{C}(\bm{\alpha}_i)$, where $\bm{\alpha}_i$ is computed via  the maximum likelihood estimate, $\bm{\alpha}_{i,t}\propto\frac{1}{t}\sum^t_{\tau=1}\mathbf{e}_{z_{i,\tau}}$, see Section \ref{sec:tf_derivation} for details. Indeed, this is similar to the TF proposal, expect it does not use the forgetting factor, $\gamma_t$, specified by the Robbins-Monro sequence, $(\gamma_t)^T_{t=1}$. In the following experiments, we use $\gamma_t=1$ for $t=1,\ldots,500$ and $\gamma_t=0.1$ otherwise. We do not use the annealing in the following experiments. To evaluate the proposal distributions, we compute the average acceptance ratio (AAR) over all $i\in(1,\ldots,N)$ at each iteration, $t$, i.e.\ $\text{AAR}_t\coloneqq\frac{1}{N}\sum^N_{i=1}\alpha(z_{i,t-1},z_{i,t})$.

\begin{figure*}
	\centering
	\input{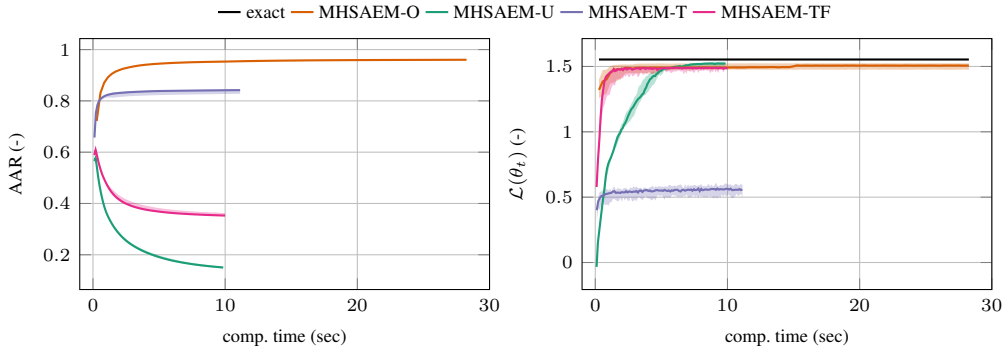}
	\caption{The average acceptance ratio, $\text{AAR}_t\coloneqq\frac{1}{N}\sum^N_{i=1}\alpha(z_{i,t-1},z_{i,t})$, (left) and the training log-likelihood, $\mathcal{L}(\theta_t)$, (right) versus the computational time (in seconds). This experiment was performed with the following settings: $(D,K,N,\omega,B, M,T)$ $=(2,10,10k,0.1,200,1,20k)$. The results are averaged over ten repetitions. The solid line is the median and the shaded area is the inter-quartile range.}
	\label{fig:ratio_and_loglik_vs_computational_time}
\end{figure*}

Figure \ref{fig:ratio_and_loglik_vs_computational_time} shows that, as one may expect, the acceptance ratio of the O-proposal converges close to one, and the corresponding log-likelihood quickly converges near to the exact value. The acceptance ratio of the U-proposal converges towards $\frac{1}{K}$, as the uniform proposal has exactly $\frac{1}{K}$ probability to sample the most representative component for a given datapoint. However, this proposal is less efficient, as can be seen from slower convergence of the corresponding log-likelihood. The TF-proposal improves the performance over the U-proposal by having a higher acceptance ratio and quicker convergence of the log-likelihood. However, the T-proposal fails to learn a sufficiently close representation of the exact posterior distribution $p_{\theta}(z|x)$, as seen from high acceptance ratio and poor log-likelihood. This result demonstrates that introducing the Robbins-Monro schedule is crucial for improving the performance of the T-proposal.

The experiments presented in Section 6.1, Section 9.1 and Section 9.2 show that the U-proposal and the TF-proposal have similar performance. This indicates that the TF-proposal looses its efficiency for high $K$. We plan to address this in future work. Nonetheless, from all these results, we observe that even the simple U-proposal achieves a substantial speed-up compared to the other algorithms in Table 1. Designing a better proposal distribution will further decrease the computational time of our MHSAEM algorithm.

\section{Sum-product-transform networks and mixtures of real NVP flows}
This section extends experiments with the SPTNs and the mixtures of real NVP flows given in Section 6.2 and Section 6.3, respectively.

\subsection{Computational cost for a fixed architecture}
In Table 2, we have compared the computational times and the test log-likelihoods for the best SPTN architectures and the best mixtures of real NVP flow models selected based on the validation log-likelihood. This experiment allows us to find the algorithm that trains the most suitable model in the shortest time. One the one hand, it reveals clear cases where the MHSAEM algorithm outperforms the SGD algorithm in terms of the test log-likelihood and the speed-up (e.g.\ \texttt{waveform}). In some situations this holds for the same architecture selected by both the SGD and MHSAEM algorithms (e.g.\ \texttt{wine}). We also see that using the SGD algorithm to fit an SPTN with $K=1024$ takes longer time than using our MHSAEM algorithm to fit an SPTN with $K=16384$, which is, however, not beneficial due to the lower values of the test log-likelihood (e.g.\ \texttt{pendigits}). On the other hand, there are situations showing no speed-up on small architectures, despite having a better log-likelihood (e.g.\ \texttt{pima-indians}). This is caused by that the implementation of the SGD algorithm is better than that of the HMSAEM algorithm.

In those lines where $K$ is substantially different for the SGD and MHSAEM algorithms, it may be difficult to recognize the maximal achievable speed-up of the MHSAEM algorithm. Therefore, we compare these algorithms also for a fixed architecture (and model) in Table \ref{tab:datasets_sptn_fixed_architecture}. For all datasets, our MHSAEM algorithm delivers a speed-up ranging approximately from 11 to 186 for SPTNs and from 1.3 to 38.6 for the mixtures of real NVP flows.

\subsection{Overfit of the mixtures of real NVP flows}
Certain models can be too complex for small and simple datasets. To consider this in our context, we present Table \ref{tab:datasets_sptn_best_architecture_all_logliks}, which extends Table 2 by additionally presenting the training and validation log-likelihoods. Here, we observe that some instances of the mixtures of real NVP flows deliver too high training log-likelihoods. If we were select these models based on their validation log-likelihoods, the resulting test log-likelihood would be rather poor. This indicates a clear overfit. The results achieved by the SPTNs are more consistent in this respect, being less prone to the model overfit. Note, too, that the mixtures of real NVP flows and the SPTNs deliver comparable results in some cases.

\section{Experiment settings}
This section provides details on the settings of experiments in Section 6 of the main paper.

\subsection{Gaussian mixture models}
The difficulty of learning GMMs heavily depends on the degree of interaction among all mixture components, hence having the ability to generate synthetic datasets with arbitrary overlap characteristics between all pairs of components is crucial for systematic evaluation of performance of learning algorithms \citesupp{naim2012convergence}. Traditional techniques usually define overlap (or separation) of components only in terms of their mean vectors and maximum eigenvalues of the covariance matrices, not accounting for their rotation and mixing weights (see \citesupp{maitra2010simulating} for a detailed treatment of the problem). We therefore use a more objective measure of the clustering cost defined by the total probability of misclassification \citesupp{melnykov2012mixsim}, which allows to generate data with a user-defined degree of maximum pairwise overlap,~$\omega$.

All the SA-variants in Table~1 (the main paper) use a minibatch of size $B$. The key quantity to reduce the number of evaluated components and/or sufficient statistics in the SSAEM, TSAEM, MCSAEM and MHSAEM algorithms is collectively denoted by $M$ (Section 5). Note that we always keep $M=\bar{M}$ in the TSAEM algorithm (see Figure~1 and~2 of the main paper for concrete numbers). We use the step-size given by $\gamma_t=1$ for $t=1,\ldots,50$ and $\gamma_t=0.05$ otherwise. In this section, to counteract the issue of attaining poor local optima, we equip \emph{all} algorithms with the anti-annealing schedule $(\beta_t)^T_{t=1}$, starting with $\beta_{1}=0.1$, reaching $\beta_{2/3T}=1.2$, and decreasing back to $\beta_{T}=1.0$, see \citesupp{naim2012convergence} for details. The initial estimates of: (i) $\mu_k$ are uniformly drawn from the unit hyper-cube, (ii) $\Sigma_k$ are fixed to unit diagonal matrix, and (iii) $\pi_k$ are uniformly drawn from the unit interval (followed by normalization).

\subsection{Sum-product transform networks}
To reduce the space of possible architectures, we restrict ourselves only to (i) the leaf nodes given by $\mathcal{N}(0, \mathbf{I})$; (ii) affine transformations fixed to the singular value decomposition, choosing the the Givens parameterization for the unitary matrices~\citesupp{pevny2020sum}; and (iii) no sharing of any type of nodes \citesupp{pevny2020sum}.

In our implementation, both the SGD and MHSAEM-U methods perform optimization of their respective objective functions---the log-likelihood 11 for SGD and the EM objective 13 for MHSAEM-U---via the use of the automatic differentiation and the ADAM optimizer \citesupp{kingma2014adam}, using $B=100$ and $T=20000$.

\subsection{Mixtures of real NVP flows}
Each real NVP-based component in the mixture model relies on (i) the translation function parameterized via multi-layer perceptron with a single hidden layer of dimension 10, using the rectified linear activation function; and (ii) the scale function parameterized via the same network except with the hyperbolic tangent activation function. We do not use the batch normalization \citesupp{dinh2017density}, and we stack two layers of the translation-scale transformation.

\section{More remarks on related work}
Various methods to decrease the computational cost of fitting large mixture models via the first three factors in $\mathcal{O}(TDNK)$ have been proposed. $T$ can be lowered by proper initialization, e.g.\ the optimal seeding \citesupp{blomer2016adaptive}; an efficient step-size schedule, e.g.\ the line-search \citesupp{xiang2020exact}; or increased estimation precision, e.g.\ the variance reduction \citesupp{chen2018stochastic}. $N$ is often reduced using the coreset methods, which approximate the original dataset by a weighted dataset such that the exact and approximate marginal likelihoods are close. The weighted variants of the variational \citesupp{feldman2011scalable, zhang2016computational,campbell2019sparse} and sampling-based \citesupp{mcgrory2014weighted} methods then process the coresets. Reducing $D$ relies on the compression of data into smaller representations via random projections \citesupp{siblini2019review,ayesha2020overview}, which is achieved in two ways: (i) each data item is projected into an individual representation \citesupp{dasgupta1999learning}; (ii) all data items are projected into an overall representation, commonly referred to as sketch \citesupp{keriven2018sketching,gribonval2020sketching}.

\section{Derivation of the TF proposal distribution}\label{sec:tf_derivation}
To ease the presentation, we assume $M=1$ in Algorithm 1. The extension to $M>1$ is straightforward. Before sampling from the proposal distribution, $q(z|\bar{z})$, we assigns $\bar{z}\coloneqq z_{i,t}$, for each $i\in I$ and $t\in(1,\ldots,T)$ in Algorithm 1. For a high enough $T$, all $x_i$s will eventually be visited, which implies that we have to learn the proposal distribution for each $i\in(1,\ldots,N)$ from the Markov chain trajectories $(z_{i,t})^T_{t=1}$.

As mentioned in the main paper, the main goal in designing an efficient proposal distribution is to bound its computational cost between that of the U and O proposals. A natural way to accomplish this is to learn $K\times K$-matrix, see, e.g.\ \citesupp{bai1975efficient}, via each Markov chain, $(z_{i,t})^T_{t=1}$, facilitating direct MC sampling from $q(\cdot|\bar{z})$---a conditional distribution resulting from the per-column normalization of the $K\times K$-matrix. Unfortunately, this would require us to store the $K\times K$-matrix for each $i\in(1,\ldots,N)$, which is very demanding even for moderate $K$ and $N$.

Therefore, to reduce the memory requirements, we break the dependence in the Markov chain, and, for each $i\in(1,\ldots,N)$, we define: $$q(z|\bar{z})\coloneqq q_{\bm{\alpha}_i}(z)\coloneqq\mathcal{C}(\bm{\alpha}_i),$$ where $\mathcal{C}(\bm{\alpha}_i)\propto\Pi^K_{k=1}\alpha^{\mathds{1}(z=k)}_{k,i}$ is the categorical distribution with the weights $\bm{\alpha}_i\coloneqq (\alpha_{1,i},\ldots,\alpha_{K,i})$. We perform the maximum likelihood estimation to find the estimate of $\bm{\alpha}_i$ at iteration $t$ as follows:
\begin{align}
\bm{\alpha}_{i,t}
&\coloneqq
\operatorname{arg\hspace{1pt}max}_{\bm{\alpha}_{i}}\mathcal{L}(\bm{\alpha}_i)
,
\nonumber
\\
\text{s. t.}&
\textstyle\sum^K_{k=1}\alpha_{k,i}=1,
\hspace{5pt}
0\leq\alpha_{k,i}\leq 1
,
\hspace{5pt}
k\in(1,\ldots,K)
,
\nonumber
\end{align}
where $\mathcal{L}(\bm{\alpha}_i)\coloneqq\Sigma^t_{\tau=1}\log q_{\bm{\alpha}_i}(z_{i,\tau})$. After the maximization, the resulting estimate is $$\bm{\alpha}_{i,t}=\frac{n_{i,t}}{\sum^K_{k=1}n_{k,i,t}},$$ where $n_{k,i,t}$ is the $k$th entry of $n_{i,t}=\Sigma^t_{\tau=1}\mathbf{e}_{z_{i,\tau}}$, and $\mathbf{e}_{k}$ is the standard basis vector (a one-hot vector) with one at $k$th position and zeros otherwise. This can be further rewritten into a recursive form: $n_{i,t}=n_{i,t-1}+\mathbf{e}_{z_{i,t}}$ or, using the Robbins-Monro step-size, $$n_{i,t}=(\bm{1}-\mathbf{e}_{z_{i,t}}\gamma_t)\odot n_{i,t-1}+\gamma_t\mathbf{e}_{z_{i,t}}.$$ $\bm{1}$ is the vector of ones, and $\odot$ is the Hadamard product.

\small
\bibliographystylesupp{aaai22}
\bibliographysupp{aaai22}

\setlength{\tabcolsep}{4pt}
\begin{table*}
	\centering
	\tiny
	\begin{tabular}{r|rrr|rrr}
		\hline\hline
		& \multicolumn{3}{c|}{Sum-product transform networks}                              & \multicolumn{3}{c}{Mixtures of real NVP flows}                                    \\\cline{2-7}
		&                   & \textbf{SGD}                 & \textbf{MHSAEM-U}             &                   & \textbf{SGD}                  & \textbf{MHSAEM-U}             \\
		\textbf{dataset}        & \textbf{speed-up} & $\mathcal{L}^{\text{test}}$  & $\mathcal{L}^{\text{test}}$   & \textbf{speed-up} & $\mathcal{L}^{\text{test}}$   & $\mathcal{L}^{\text{test}}$   \\\hline
		breast-cancer-wisconsin & 16.72             & -24.01                       & \color{blue}{\textbf{0.2}}    & 2.62              & -112.63                       & \color{blue}{\textbf{-39.31}} \\
		cardiotocography        & 17.93             & \color{blue}{\textbf{45.81}} & 31.04                         & 27.45             & 45.99                         & \color{blue}{\textbf{56.08}}  \\
		magic-telescope         & 24.21             & \color{blue}{\textbf{-4.41}} & -5.21                         & 20.41             & -6.01                         & \color{blue}{\textbf{-4.96}}  \\
		pendigits               & 15.5              & \color{blue}{\textbf{-0.58}} & -5.96                         & 17.82             & \color{blue}{\textbf{-1.78}}  & -3.7                          \\
		pima-indians            & 17.69             & -20.18                       & \color{blue}{\textbf{-8.3}}   & 1.35              & -20.09                        & \color{blue}{\textbf{-16.33}} \\
		wall-following-robot    & 16.96             & \color{blue}{\textbf{-6.05}} & -16.98                        & 22.21             & \color{blue}{\textbf{-14.26}} & -17.56                        \\
		waveform-1              & 101.37            & -39.16                       & \color{blue}{\textbf{-23.94}} & 19.09             & \color{blue}{\textbf{-39.06}} & -40.41                        \\
		waveform-2              & 92.73             & -38.87                       & \color{blue}{\textbf{-23.95}} & 20.32             & \color{blue}{\textbf{-39.42}} & -40.79                        \\
		yeast                   & 12.41             & 4.29                         & \color{blue}{\textbf{5.18}}   & 14.49             & 6.61                          & \color{blue}{\textbf{9.59}}   \\
		ecoli                   & 29.45             & -14.02                       & \color{blue}{\textbf{-0.22}}  & 2.15              & -11.37                        & \color{blue}{\textbf{-10.64}} \\
		ionosphere              & 13.78             & -30.25                       & \color{blue}{\textbf{-7.35}}  & 2.74              & -87.01                        & \color{blue}{\textbf{-42.75}} \\
		iris                    & 52.62             & -12.5                        & \color{blue}{\textbf{-1.76}}  & 1.45              & -16.34                        & \color{blue}{\textbf{-9.45}}  \\
		page-blocks             & 16.68             & \color{blue}{\textbf{11.98}} & 6.84                          & 26.04             & 17.13                         & \color{blue}{\textbf{17.27}}  \\
		parkinsons              & 16.64             & -35.75                       & \color{blue}{\textbf{-1.15}}  & 3.22              & -566.58                       & \color{blue}{\textbf{-33.0}}  \\
		sonar                   & 11.5              & -98.43                       & \color{blue}{\textbf{-88.33}} & 2.52              & -622.2                        & \color{blue}{\textbf{-88.81}} \\
		statlog-satimage        & 14.52             & 3.33                         & \color{blue}{\textbf{3.74}}   & 25.31             & \color{blue}{\textbf{-7.38}}  & -17.89                        \\
		statlog-segment         & 16.35             & \color{blue}{\textbf{40.81}} & 25.73                         & 38.68             & 35.84                         & \color{blue}{\textbf{36.8}}   \\
		statlog-vehicle         & 23.43             & -22.52                       & \color{blue}{\textbf{-5.45}}  & 33.71             & -31.21                        & \color{blue}{\textbf{-26.43}} \\
		wine                    & 186.25            & -26.53                       & \color{blue}{\textbf{-13.5}}  & 2.05              & -171.58                       & \color{blue}{\textbf{-25.57}} \\\hline
		rank                    &                   & 1.68                         & \color{blue}{\textbf{1.32}}   &                   & 1.74                          & \color{blue}{\textbf{1.26}}   \\\hline\hline
	\end{tabular}
	\caption{The speed-up and test log-likelihood, $\mathcal{L}^{\text{test}}$, for the SGD and MHSAEM-U algorithms. The test log-likelihood (higher is better) is computed for the SPTN with $K=1024$ and the mixture of real NVP flows with $K=128$. The speed-up is computed as the ratio of MHSAEM-U to SGD, i.e.\ their time to reach 95\% of the training log-likelihood. The results are averaged over five repetitions with different initial conditions. Then, the higher test log-likelihood is highlighted with bold blue. The average rank is computed as the standard competition (``1224'') ranking \protect\citesupp{demvsar2006statistical} on each dataset (lower is better).}
	\label{tab:datasets_sptn_fixed_architecture}
\end{table*}

\setlength{\tabcolsep}{4pt}
\begin{table*}
	\centering
	\tiny
	\begin{tabular}{r|rrrrrr|rrrrrr}
		\hline\hline
		& \multicolumn{6}{c|}{Sum-product transform networks} & \multicolumn{6}{c}{Mixtures of real NVP flows}\\\cline{2-13}
		& \multicolumn{2}{c}{$\mathcal{L}(\theta_{t_{95}})$} & \multicolumn{2}{c}{$\mathcal{L}^{\text{test}}$} & \multicolumn{2}{c|}{$\mathcal{L}^{\text{val}}$} & \multicolumn{2}{c}{$\mathcal{L}(\theta_{t_{95}})$} & \multicolumn{2}{c}{$\mathcal{L}^{\text{test}}$} & \multicolumn{2}{c}{$\mathcal{L}^{\text{val}}$}\\
		\textbf{dataset} & \textbf{SGD} & \textbf{MHSAEM-U} & \textbf{SGD} & \textbf{MHSAEM-U} & \textbf{SGD} & \textbf{MHSAEM-U} & \textbf{SGD} & \textbf{MHSAEM-U} & \textbf{SGD} & \textbf{MHSAEM-U} & \textbf{SGD} & \textbf{MHSAEM-U} \\\hline
		breast-cancer-wisconsin & \color{blue}{\textbf{11.88}} & 5.47 & -4.66 & \color{blue}{\textbf{1.43}} & -5.89 & \color{blue}{\textbf{0.33}} & 21.19 & \color{blue}{\textbf{46.97}} & -99.85 & \color{blue}{\textbf{-39.31}} & -70.33 & \color{blue}{\textbf{-39.21}} \\
		cardiotocography & \color{blue}{\textbf{59.72}} & 30.36 & \color{blue}{\textbf{59.52}} & 31.04 & \color{blue}{\textbf{59.09}} & 30.06 & 58.21 & \color{blue}{\textbf{62.95}} & 54.34 & \color{blue}{\textbf{56.08}} & 52.7 & \color{blue}{\textbf{54.35}} \\
		magic-telescope & \color{blue}{\textbf{-3.08}} & -5.24 & \color{blue}{\textbf{-3.65}} & -5.03 & \color{blue}{\textbf{-3.75}} & -5.11 & \color{blue}{\textbf{-3.79}} & -4.07 & \color{blue}{\textbf{-3.97}} & -4.22 & \color{blue}{\textbf{-4.1}} & -4.31 \\
		pendigits & \color{blue}{\textbf{4.89}} & -4.21 & \color{blue}{\textbf{0.88}} & -4.86 & \color{blue}{\textbf{0.77}} & -4.77 & \color{blue}{\textbf{5.09}} & 3.94 & \color{blue}{\textbf{1.46}} & 0.48 & \color{blue}{\textbf{1.66}} & 0.77 \\
		pima-indians & \color{blue}{\textbf{-7.22}} & -7.4 & -8.54 & \color{blue}{\textbf{-7.62}} & -8.63 & \color{blue}{\textbf{-7.54}} & 1.97 & \color{blue}{\textbf{11.52}} & -20.09 & \color{blue}{\textbf{-16.33}} & -21.48 & \color{blue}{\textbf{-16.61}} \\
		wall-following-robot & \color{blue}{\textbf{15.9}} & -6.75 & \color{blue}{\textbf{1.84}} & -11.3 & \color{blue}{\textbf{-0.16}} & -11.36 & \color{blue}{\textbf{26.53}} & 22.95 & \color{blue}{\textbf{-14.26}} & -17.56 & \color{blue}{\textbf{-15.99}} & -19.11 \\
		waveform-1 & \color{blue}{\textbf{-22.34}} & -24.4 & -26.14 & \color{blue}{\textbf{-23.91}} & -25.93 & \color{blue}{\textbf{-23.86}} & \color{blue}{\textbf{-20.94}} & -21.01 & -34.12 & \color{blue}{\textbf{-33.42}} & -34.31 & \color{blue}{\textbf{-33.6}} \\
		waveform-2 & \color{blue}{\textbf{-22.17}} & -24.4 & -26.21 & \color{blue}{\textbf{-23.91}} & -25.87 & \color{blue}{\textbf{-23.86}} & \color{blue}{\textbf{-20.87}} & -21.03 & -34.15 & \color{blue}{\textbf{-33.64}} & -34.02 & \color{blue}{\textbf{-33.54}} \\
		yeast & \color{blue}{\textbf{13.63}} & 5.59 & \color{blue}{\textbf{10.26}} & 5.18 & \color{blue}{\textbf{10.81}} & 5.79 & 10.26 & \color{blue}{\textbf{14.0}} & 6.61 & \color{blue}{\textbf{9.59}} & 8.45 & \color{blue}{\textbf{10.79}} \\
		ecoli & \color{blue}{\textbf{6.3}} & 0.65 & -5.5 & \color{blue}{\textbf{-0.22}} & -6.36 & \color{blue}{\textbf{-0.83}} & 5.55 & \color{blue}{\textbf{10.78}} & -11.37 & \color{blue}{\textbf{-10.64}} & -46.87 & \color{blue}{\textbf{-11.66}} \\
		ionosphere & \color{blue}{\textbf{33.58}} & 6.87 & -20.27 & \color{blue}{\textbf{-5.93}} & -12.39 & \color{blue}{\textbf{-1.87}} & 43.3 & \color{blue}{\textbf{74.58}} & -87.01 & \color{blue}{\textbf{-42.75}} & -102.24 & \color{blue}{\textbf{-41.88}} \\
		iris & \color{blue}{\textbf{4.7}} & -1.58 & -10.65 & \color{blue}{\textbf{-1.49}} & -12.11 & \color{blue}{\textbf{-1.84}} & 1.17 & \color{blue}{\textbf{7.64}} & -16.34 & \color{blue}{\textbf{-9.21}} & -13.88 & \color{blue}{\textbf{-9.05}} \\
		page-blocks & \color{blue}{\textbf{12.76}} & 6.73 & \color{blue}{\textbf{12.21}} & 6.84 & \color{blue}{\textbf{12.36}} & 6.93 & 17.67 & \color{blue}{\textbf{18.62}} & 17.13 & \color{blue}{\textbf{17.94}} & 17.42 & \color{blue}{\textbf{18.01}} \\
		parkinsons & \color{blue}{\textbf{27.23}} & 5.3 & -21.85 & \color{blue}{\textbf{0.5}} & -23.61 & \color{blue}{\textbf{-0.25}} & 33.97 & \color{blue}{\textbf{50.33}} & -566.58 & \color{blue}{\textbf{-33.31}} & -552.33 & \color{blue}{\textbf{-33.46}} \\
		sonar & \color{blue}{\textbf{158.64}} & -16.24 & -95.39 & \color{blue}{\textbf{-69.29}} & -96.02 & \color{blue}{\textbf{-72.51}} & 78.05 & \color{blue}{\textbf{142.83}} & -622.2 & \color{blue}{\textbf{-88.81}} & -371.82 & \color{blue}{\textbf{-87.63}} \\
		statlog-segment & \color{blue}{\textbf{51.73}} & 28.01 & \color{blue}{\textbf{47.35}} & 26.53 & \color{blue}{\textbf{44.79}} & 26.3 & 47.46 & \color{blue}{\textbf{50.17}} & 35.84 & \color{blue}{\textbf{42.04}} & 34.8 & \color{blue}{\textbf{41.4}} \\
		statlog-vehicle & \color{blue}{\textbf{3.43}} & -4.0 & \color{blue}{\textbf{-4.25}} & -5.45 & \color{blue}{\textbf{-4.83}} & -5.31 & 14.79 & \color{blue}{\textbf{22.51}} & -31.34 & \color{blue}{\textbf{-26.43}} & -30.23 & \color{blue}{\textbf{-25.92}} \\
		wine & \color{blue}{\textbf{43.01}} & -12.39 & -25.99 & \color{blue}{\textbf{-13.27}} & -26.17 & \color{blue}{\textbf{-12.95}} & 19.21 & \color{blue}{\textbf{34.77}} & -171.58 & \color{blue}{\textbf{-25.57}} & -262.69 & \color{blue}{\textbf{-25.03}} \\
		rank & \color{blue}{\textbf{1.0}} & 2.0 & 1.56 & \color{blue}{\textbf{1.44}} & 1.56 & \color{blue}{\textbf{1.44}} & 1.72 & \color{blue}{\textbf{1.28}} & 1.83 & \color{blue}{\textbf{1.17}} & 1.83 & \color{blue}{\textbf{1.17}} \\\hline\hline
	\end{tabular}
	\caption{95\% of the maximal training log-likelihood, $\mathcal{L}(\theta_{t_{95}})$, validation log-likelihood, $\mathcal{L}^{\text{val}}$, and test log-likelihood, $\mathcal{L}^{\text{test}}$, for the SGD and MHSAEM-U algorithms (higher is better). The lines in this table correspond to those in Table 2, except we add the training and validation log-likelihoods and do not repeat the speed-up and the number of components, $K$ (see Table 2 for completeness). The results are averaged over five repetitions with different initial conditions. Then, the higher log-likelihood is highlighted with bold blue. The average rank is computed as the standard competition (``1224'') ranking \protect\citesupp{demvsar2006statistical} on each dataset (lower is better).}
	\label{tab:datasets_sptn_best_architecture_all_logliks}
\end{table*}

\end{multicols}

\end{document}